\documentclass[letterpaper, 10 pt, conference]{ieeeconf}  
\IEEEoverridecommandlockouts
\usepackage{threeparttable}
\usepackage{cite}
\usepackage{amsmath,amssymb,amsfonts}
\usepackage{graphicx}
\usepackage{textcomp}
\usepackage{color}
\usepackage{xcolor}
\usepackage{colortbl}
\usepackage{subfigure}
\usepackage{multirow}
\usepackage{multicol}
\usepackage{makecell}
\usepackage{bm}
\usepackage{algorithm}
\usepackage{algpseudocode}
\usepackage{booktabs} 
\usepackage[left=48pt, right=48pt, top=57pt, bottom=43pt]{geometry}
\usepackage[colorlinks,bookmarks,bookmarksopen,bookmarksnumbered,linktocpage,hyperindex,citecolor=blue, linkcolor=blue, urlcolor=blue]{hyperref}

\newcommand{\Reffig}[1]{Fig.~\ref{#1}}

\newcommand{\Reftab}[1]{Table~\ref{#1}}

\begin{document}
\title{
        Learning from Mistakes: Loss-Aware Memory Enhanced Continual Learning for LiDAR Place Recognition 
}


\author{Xufei Wang$^{1}$, Junqiao Zhao$^{*, 1, 2, 3}$, Siyue Tao$^{2}$, Qiwen Gu$^{2}$, Wonbong Kim$^{2}$, Tiantian Feng$^{4}$ 
\thanks{This work is supported by XXX. \emph{(Corresponding Author: Junqiao Zhao.)}}
\thanks{$^{1}$Xufei Wang, Junqiao Zhao are with the Shanghai Research Institute for Intelligent Autonomous System, Tongji University, Shanghai, China
{\tt\footnotesize (Corresponding Author: zhaojunqiao@tongji.edu.cn).}}
\thanks{$^{2}$Siyue Tao, Qiwen Gu, Wonbong Kim, and Junqiao Zhao are with the Department of Computer Science and Technology, School of Electronics and Information Engineering, Tongji University, Shanghai, China, and the MOE Key Lab of Embedded System and Service Computing, Tongji University, Shanghai, China}
\thanks{$^{3}$Institute of Intelligent Vehicles, Tongji University, Shanghai, China}
\thanks{$^{4}$College of Surveying and Geo-Informatics, Tongji University, Shanghai, China}
}

\maketitle

\begin{abstract}
LiDAR place recognition plays a crucial role in SLAM, robot navigation, and autonomous driving. However, existing LiDAR place recognition methods often struggle to adapt to new environments without forgetting previously learned knowledge, a challenge widely known as catastrophic forgetting. To address this issue, we propose \textbf{KDF+}, a novel continual learning framework for LiDAR place recognition that extends the KDF paradigm with a loss-aware sampling strategy and a rehearsal enhancement mechanism.
The proposed sampling strategy estimates the learning difficulty of each sample via its loss value and selects samples for replay according to their estimated difficulty. Harder samples, which tend to encode more discriminative information, are sampled with higher probability while maintaining distributional coverage across the dataset. In addition, the rehearsal enhancement mechanism encourages memory samples to be further refined during new-task training by slightly reducing their loss relative to previous tasks, thereby reinforcing long-term knowledge retention.
Extensive experiments across multiple benchmarks demonstrate that KDF+ consistently outperforms existing continual learning methods and can be seamlessly integrated into state-of-the-art continual learning for LiDAR place recognition frameworks to yield significant and stable performance gains.
The code will be available at \url{https://github.com/repo/KDF-plus}.

\end{abstract}

\section{INTRODUCTION}
\label{sec:introduction} 
LiDAR Place Recognition (LPR)\cite{yin2024general,shi2023lidar} aims to identify revisited locations by retrieving the most similar scan from a map database given the current LiDAR observation. It plays a crucial role in robotics and autonomous navigation, serving as the foundation for reliable localization and loop closure detection in large-scale and long-term mapping scenarios.

Recently, a wide range of neural network-based LPR methods\cite{uy2018pointnetvlad,komorowski2021minkloc3d,xia2023casspr} have achieved remarkable performance in static or closed-set environments. However, these methods tend to degrade significantly when exposed to continuously changing environments. In particular, sequential learning scenarios often induce catastrophic forgetting\cite{wang2024comprehensive}, where the model struggles to retain previously learned knowledge.

To address this challenge, several studies\cite{knights2022irosincloud,cui2023ralccl,yin2023bioslam,liu2024micl,wang2025ranking} have incorporated continual learning techniques\cite{wang2024comprehensive} into the LPR domain, giving rise to the emerging research area of Continual Learning for LiDAR Place Recognition (CL-LPR). Representative works such as InCloud\cite{knights2022irosincloud}, CCL\cite{cui2023ralccl}, MICL\cite{liu2024micl}, and KDF\cite{wang2025ranking} explore various knowledge distillation strategies to mitigate forgetting while preserving model adaptability.

\begin{figure}[t]
        \centering
        \includegraphics[width=\columnwidth]{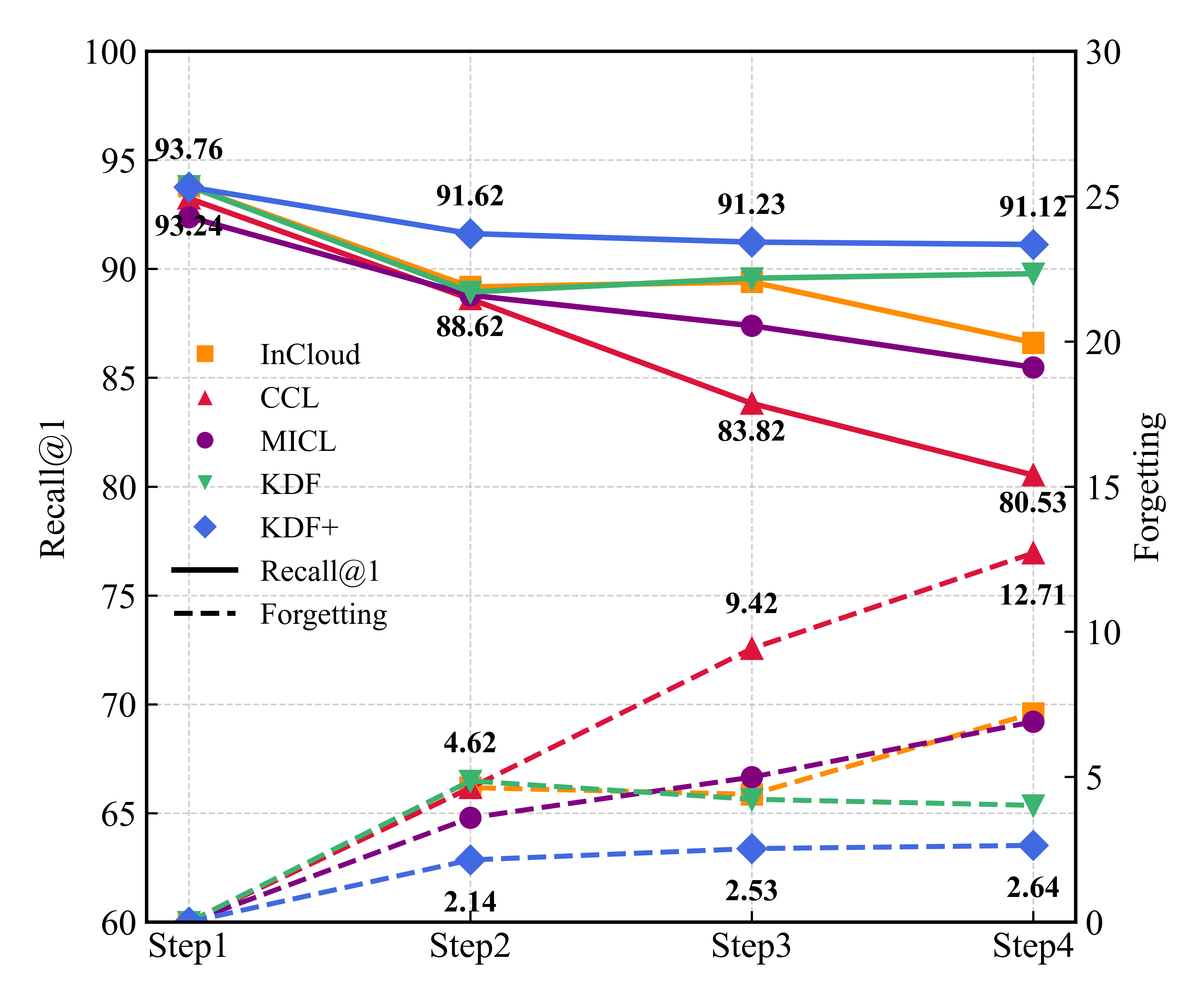}
        \caption{Comparison of different methods in mitigating catastrophic forgetting for LiDAR place recognition (LPR). At each step, the trained MinkLoc3D\cite{komorowski2021minkloc3d} model is evaluated on the Oxford\cite{maddern20171} dataset, and both Recall@1 and Forgetting scores are reported. Solid lines represent Recall@1, while dashed lines indicate the Forgetting score. Higher Recall@1 values correspond to better place recognition performance, whereas lower Forgetting scores reflect stronger resistance to catastrophic forgetting.As shown in the figure, baseline methods such as InCloud\cite{knights2022irosincloud}, CCL\cite{cui2023ralccl}, and MICL\cite{liu2024micl} experience substantial drops in Recall@1 and sharp increases in Forgetting as new tasks are learned. The previous method KDF\cite{wang2025ranking} provides moderate improvements. In contrast, our proposed KDF+ consistently maintains the highest Recall@1 and achieves the lowest Forgetting score across all steps, demonstrating its superior capability in mitigating catastrophic forgetting while preserving recognition performance on previously learned tasks.
        }\label{fig:teaser}
        
\end{figure}

Most existing methods\cite{knights2022irosincloud,cui2023ralccl,yin2023bioslam,liu2024micl,wang2025ranking} rely on maintaining a memory (replay) buffer to enhance knowledge retention and mitigate forgetting. As summarized in \Reftab{tab:memory_replay_comparison}, prior continual learning approaches differ widely in how these replay buffers are constructed and utilized. While random sampling is commonly employed—likely due to its simplicity and ease of implementation—it often fails to preserve the most informative samples for long-term retention and ignores the model’s perception of sample difficulty. 
Furthermore, these approaches also vary in how replayed samples are incorporated during training. Several methods directly mix memory samples with new-task data for joint optimization and use them solely for knowledge distillation, a straightforward strategy but one that lacks an optimized or principled replay mechanism. Discrepancies in memory management policies across different methods further contribute to inconsistent performance and inefficiencies in maintaining previously acquired knowledge.

To address these limitations, we propose a new continual learning framework for LiDAR place recognition, termed \textbf{KDF+}, which extends the original KDF framework with a \emph{loss-aware sampling mechanism} and a \emph{rehearsal enhancement mechanism}. As illustrated in \Reffig{fig:teaser}, the enhanced KDF+ framework effectively mitigates forgetting in continual LPR settings. Unlike prior methods, our sampling strategy estimates the difficulty of each sample based on its training loss, following the intuition that harder samples contain richer supervisory signals. Consequently, samples are probabilistically selected for memory replay according to their estimated difficulty, yielding a more informative and representative buffer.
In addition, our rehearsal enhancement mechanism promotes further optimization of memory samples during new-task training by encouraging their loss to be slightly reduced relative to previous tasks. This design reinforces long-term knowledge retention and enables memory samples to be continually improved throughout sequential learning.

\begin{table}[t!]
        \centering
        \caption{Comparison of existing CL-LPR methods.}
        \label{tab:memory_replay_comparison}
        \setlength{\tabcolsep}{3pt}
        \normalsize
        \resizebox{\columnwidth}{!}{
        \begin{threeparttable}
        \begin{tabular}{lcccc}
        \toprule
        Method & Knowledge Distillation & Data Sampling & Replay Strategy & Memory Update \\
        \midrule
        InCloud\cite{knights2022irosincloud} & Structure-aware & Random & Mix & Max replacement \\
        CCL\cite{cui2023ralccl}     & Distribution-based & Random & Sample ER & Max replacement \\
        MICL\cite{liu2024micl}    & Mutual Information-based & Random & Pairs ER+Mix & Balanced \\
        KDF\cite{wang2025ranking}     & Ranking-aware & Random & Mix & Max replacement \\
        \bottomrule
        \end{tabular}
        \end{threeparttable}
        }
\end{table}

In summary, this paper makes the following key contributions:
\begin{itemize}
    \item We propose a \emph{loss-aware memory sampling strategy} that adaptively selects informative and challenging samples for replay based on their estimated difficulty.
    \item We introduce a \emph{rehearsal enhancement loss} that facilitates more effective refinement of memory samples during continual updates, thereby reinforcing long-term knowledge retention.
    \item Extensive experiments demonstrate that the proposed KDF+ framework achieves superior performance compared with existing methods, and can be seamlessly combined with state-of-the-art continual learning baselines to yield consistent and significant improvements.
\end{itemize}

\section{RELATED WORKS}
\label{sec:related works}
\subsection{LiDAR Place Recognition}
LiDAR Place Recognition has witnessed significant progress in recent years, largely driven by advancements in deep learning for point cloud processing. Early methods such as PointNetVLAD\cite{uy2018pointnetvlad} combined PointNet\cite{qi2017pointnet} with NetVLAD\cite{arandjelovic2016netvlad} to learn global descriptors directly from raw point clouds. Subsequent studies\cite{komorowski2021minkloc3d,xia2023casspr, komorowski2022improving, vidanapathirana2022logg3d, xu2023transloc3d,kong2020semantic,vidanapathirana2021locus,li2024efficient, li2022rinet, zhao2022spherevlad++,xia2024text2loc, shang2024mambaplace} explored diverse architectures and loss functions to improve descriptor robustness. For example, MinkLoc3D\cite{komorowski2021minkloc3d} introduced sparse 3D convolutions to efficiently handle large-scale point clouds, while CASSRP\cite{xia2023casspr} employed cross-attention mechanisms to capture richer contextual relationships within the data.

Despite these advancements, most existing LPR methods are developed under static or closed-set assumptions and thus struggle to cope with continuously evolving environments. In real-world deployments, models must adapt to new conditions while retaining previously acquired knowledge, yet standard LPR approaches often suffer from severe performance degradation due to forgetting. These practical challenges have motivated growing interest in incorporating continual learning techniques into LPR systems, giving rise to Continual Learning for LiDAR Place Recognition as an emerging and important research direction.

\subsection{Continual Learning}
Continual Learning\cite{wang2024comprehensive}, also known as Lifelong Learning, aims to enable an intelligent agent to acquire new knowledge from a non-stationary data stream while retaining previously learned capabilities. To mitigate catastrophic forgetting, existing continual learning approaches can be broadly categorized into three major families: replay-based, regularization-based, and architecture-based methods.

\textbf{Replay-based (or rehearsal-based) methods:} These approaches store a subset of samples from previous tasks and interleave them with new-task data during training. The simplest and most widely adopted strategy is Experience Replay (ER)\cite{rolnick2019experience}, which maintains a memory buffer of past samples and replays them jointly with new data (e.g., ER\cite{rolnick2019experience}, A-GEM\cite{chaudhry2019efficient}). An alternative direction is Generative Replay, where a generative model is trained to synthesize samples from old tasks, thereby reducing storage and privacy concerns (e.g., DGR\cite{shin2017continual}).

\textbf{Regularization-based methods:} These methods incorporate additional regularization terms into the loss function to constrain updates to parameters that are important for previously learned tasks. Such methods are generally divided into: (1) \emph{Weight regularization}, which estimates parameter importance and penalizes substantial changes to critical weights (e.g., EWC\cite{serra2018overcoming}, SI\cite{zenke2017continual}); and (2) \emph{Function regularization}, which leverages techniques such as knowledge distillation to ensure the new model’s outputs or features remain consistent with those of the old model (e.g., LwF\cite{li2017learning}).

\textbf{Architecture-based methods:} These approaches\cite{mallya2018packnet, rusu2016progressive} mitigate forgetting by dynamically expanding or adapting the network architecture, often allocating task-specific parameters or modules to preserve prior knowledge.

\subsection{Continual Learning for LiDAR Place Recognition}
Although continual learning has been extensively studied in image classification, only a few studies\cite{knights2022irosincloud,cui2023ralccl,yin2023bioslam,liu2024micl,wang2025ranking} have explored its application to LPR. These methods primarily rely on two strategies: knowledge distillation and sample replay.
Early attempts such as InCloud\cite{knights2022irosincloud} propose a structure-aware distillation approach that preserves higher-order embedding relationships and achieves incremental adaptation across multiple large-scale LiDAR benchmarks. CCL\cite{cui2023ralccl} introduces a continual contrastive learning paradigm with distribution-based distillation to learn transferable place representations. MICL\cite{liu2024micl} presents a mutual-information guided continual learning framework that preserves domain-shared information through a mutual-information distillation loss. KDF\cite{wang2025ranking} further advances this line of work by introducing a ranking-aware knowledge distillation and fusion framework, in which embedding ranking relations are explicitly maintained via a ranking-aware loss, and a knowledge fusion module integrates old and new model knowledge for improved LPR performance.

\begin{figure*}[t!]
        \centering
        \includegraphics[width=0.95\textwidth]{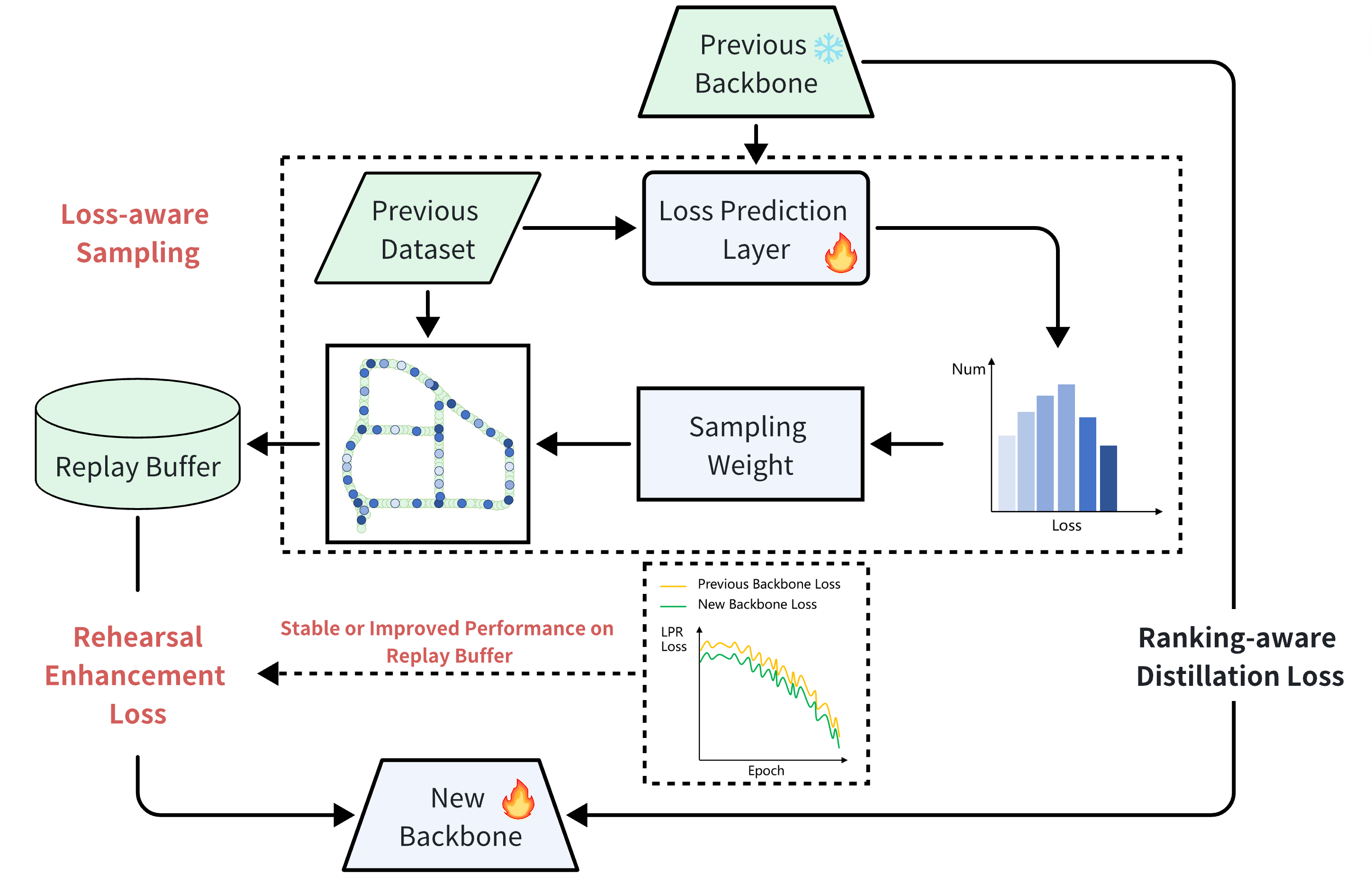}
        \caption{Overview of the proposed KDF+ CL-LPR framework. The framework incorporates two key components: (1) \textbf{Loss-Aware Sampling}: A loss prediction layer estimates the difficulty or importance of samples from the previous dataset and produces a non-uniform sampling weight distribution (illustrated by the histogram). This enables prioritized selection of informative samples (e.g., LiDAR submaps) to populate the replay buffer, as indicated by the weighted samples. (2) \textbf{Rehearsal Enhancement Loss}: Applied during rehearsal when learning a new task, this loss encourages memory samples to further improve. As shown by the loss curves, it helps maintain performance on the previous task (yellow line) while the backbone adapts to the new task (green line), effectively reducing forgetting. In addition, the framework leverages ranking-aware distillation\cite{wang2025ranking} to transfer knowledge from the previous backbone to the newly updated backbone.
        }\label{fig:pipeline}
\end{figure*}

All the above methods employ a random sampling strategy when constructing the memory buffer, without exploring more advanced or adaptive mechanisms for selecting informative samples. In addition, the stored memory samples are used solely for knowledge distillation to preserve past knowledge, with limited investigation into how these samples could be more effectively leveraged during continual updates.

\section{Preliminaries}
\label{sec:problem formulation}
Before introducing our method, we first define the LiDAR Place Recognition task and formulate the problem of continual learning for LPR.

\subsection{LiDAR Place Recognition}
LiDAR Place Recognition aims to identify previously visited locations using 3D point cloud data captured by LiDAR sensors. Given a query point cloud $P_q$, the objective is to retrieve the most similar point cloud from a database $\mathcal{D} = \{P_1, P_2, \dots, P_N\}$, where $N$ denotes the number of stored point clouds. 
To accomplish this, LPR methods typically extract global descriptors from point clouds using a feature extraction network $\Theta(\cdot)$. The similarity between the query descriptor and each database descriptor is then computed using a distance metric, such as Euclidean distance or cosine similarity. The retrieved point clouds are subsequently ranked based on their similarity scores, and the top-ranked candidates are regarded as potential matches for place recognition.

\subsection{Continual Learning for LiDAR Place Recognition}
In this paper, we formulate the continual learning problem for LPR as CL-LPR. The objective is to enable the model to acquire strong LPR capabilities in newly encountered environments while maintaining its performance on previously visited ones. Specifically, a pretrained LPR model is sequentially exposed to data from $T$ distinct domains $\{D_1, D_2, \dots, D_T\}$, each characterized by variations in environment, sensor configuration, or temporal conditions.

During each training step $t$, the model has access only to the data from the current domain $D_t$ and a predefined replay buffer (memory) $M$. The memory $M$ stores a limited number $S$ of representative samples from previously encountered domains and is incrementally updated as new domains are introduced. Consequently, the model parameters $\Theta_t$ and the replay memory $M_t$ are updated according to:
\begin{equation}
\{M_t, \Theta_t\} \leftarrow \{D_t, M_{t-1}, \Theta_{t-1}\}.
\end{equation}

Existing CL-LPR methods typically optimize a weighted combination of the task loss and a distillation loss, formulated as:
\begin{equation}
    \mathcal{L}_{\text{total}} = \mathcal{L}_{\text{PR}} + \lambda \cdot \mathcal{L}_{\text{KD}},
\end{equation}
where $\mathcal{L}_{\text{PR}}$ denotes the place recognition task loss, commonly implemented using a triplet margin loss or other contrastive loss. $\mathcal{L}_{\text{KD}}$ represents the knowledge distillation loss proposed by different methods (summarized in Table~\ref{tab:memory_replay_comparison}), and $\lambda$ is a relaxation coefficient\cite{knights2022irosincloud}.

\section{METHODOLOGY}
\label{sec:methodology}
In this section, we present the proposed continual learning framework \textbf{KDF+} for LiDAR Place Recognition. As illustrated in Figure~\ref{fig:pipeline}, the framework consists of two key components: (1) a \emph{loss-aware sampling strategy} that prioritizes the selection of informative samples for replay based on their estimated learning difficulty, and (2) a \emph{rehearsal enhancement mechanism} that encourages the model to further optimize memory samples during new-task training, thereby reinforcing long-term knowledge retention.

\subsection{Loss-Aware Sampling}

The proposed loss-aware sampling strategy addresses the problem of selecting which samples from the current dataset should be added to the replay buffer as exemplars. After multiple rounds of training, the model naturally develops an internal notion of sample difficulty, which is reflected in the loss values produced during optimization. Motivated by this observation, we introduce a lightweight loss prediction module that operates alongside the main LPR model.

As shown in Figure~\ref{fig:pipeline}, the loss prediction layer is placed after the GEM\cite{radenovic2018fine} module of the backbone network and works jointly with the entire model. It consists of a simple MLP with a Linear--BN--ReLU--Linear architecture. Given the feature representation of each point cloud, the module predicts its corresponding loss value. During training, while the original LPR model is optimized using the place recognition loss, the loss prediction layer is trained jointly under the supervision of a mean squared error (MSE) objective:
\begin{equation}
    \mathcal{L} = \mathrm{MSE}(\mathcal{L}_{\text{PR}}, \mathcal{L}_{\text{Pre}}),
\label{eq:mse}
\end{equation}
where $\mathcal{L}_{\text{Pre}}$ denotes the predicted loss.

The predicted loss values serve as an indicator of sample difficulty during the memory selection process. For each sample, a higher predicted loss typically corresponds to a more challenging case for the model, while lower predicted losses indicate easier samples that may contribute less to long-term retention. We therefore incorporate these predicted loss values as sampling weights to perform weighted random sampling. This maintains overall sampling diversity while biasing selection toward more informative (i.e., harder) samples. The sampling probability for each sample $i$ is defined as:
\begin{equation}
    p_i = \frac{\mathcal{L}^i_{\text{Pre}}}{\sum_{j} \mathcal{L}^j_{\text{Pre}}},
\end{equation}
where $\mathcal{L}^i_{\text{Pre}}$ denotes the predicted loss for sample $i$. 

By assigning higher sampling probabilities to harder samples, the model preferentially selects challenging point clouds to populate the replay buffer, enabling more effective knowledge retention during continual learning.

\subsection{Rehearsal Enhancement}

The motivation behind the rehearsal enhancement mechanism is inspired by human learning, where recalling and revisiting previously encountered experiences helps reinforce old knowledge while acquiring new information. Following this intuition, we design a rehearsal enhancement mechanism that encourages memory samples to be further refined when the model is trained in new environments.

Concretely, we expect the place recognition (PR) loss of each memory sample in a new domain to be slightly lower than its corresponding loss in the previous domain, indicating improved representation quality over time. To enforce this behavior, we introduce a rehearsal loss defined as:
\begin{equation}
    \mathcal{L}_{\text{Rehearsal}} = \max(0,\, m - (\mathcal{L}_{\text{PR}}^{old} - \mathcal{L}_{\text{PR}})),
\end{equation}
where $m$ is a margin parameter, $\mathcal{L}_{\text{PR}}^{old}$ denotes the PR loss of the memory sample in the previous domain, and $\mathcal{L}_{\text{PR}}$ represents its PR loss in the current domain.

This formulation encourages the model to continually improve its performance on previously learned samples as it adapts to new environments. By promoting progressive refinement of memory representations, the rehearsal enhancement mechanism helps reduce the impact of catastrophic forgetting and strengthens long-term knowledge retention.

\subsection{Memory Management}

\subsubsection{Equal Domain Strategy}
After training on the first domain, we initialize the replay memory by selecting a set of representative samples using the proposed loss-aware sampling strategy, subject to the predefined memory capacity $S$. After training on each subsequent domain, we follow the equal domain strategy adopted in MICL~\cite{liu2024micl} to maintain balanced memory allocation, ensuring that each domain occupies an equal number of samples, i.e., $S/T$, where $S$ is the total memory size and $T$ is the number of domains currently stored in memory.

This strategy determines how many samples should be retained or replaced for each domain. During memory updates, samples to be removed are selected via random sampling within each domain's subset, whereas new samples from the latest domain are added using the proposed loss-aware sampling approach. This maintains diversity while ensuring that informative and challenging samples are preserved.

\subsubsection{Experience Replay}
During training on a new domain, we employ experience replay\cite{rolnick2019experience} to interleave samples from the replay buffer with data from the current domain. Specifically, in each training iteration, we sample one mini-batch from the current domain and one mini-batch from the replay buffer. Each mini-batch consists of paired samples, where each pair includes an anchor and a positive example.

These two mini-batches are then combined to form a mixed batch, which is used to compute the overall loss for backpropagation. This replay mechanism enables the model to simultaneously learn from new experiences while reinforcing previously acquired knowledge, thereby effectively mitigating catastrophic forgetting.

\subsection{Overall Objective}

Most existing CL-LPR approaches primarily focus on designing distillation-based losses. In our framework, we extend the conventional loss formulation—typically consisting of a task loss and a distillation loss—by incorporating the proposed rehearsal loss. The overall training objective is defined as:
\begin{equation}
    \mathcal{L}_{\text{total}} = 
    \mathcal{L}_{\text{PR}} 
    + \lambda \cdot \mathcal{L}_{\text{KD}} 
    + \omega \cdot \mathcal{L}_{\text{Rehearsal}},
\end{equation}
where $\omega$ is a weighting coefficient that controls the contribution of the rehearsal loss. 

In this work, the place recognition task loss $\mathcal{L}_{\text{PR}}$ is implemented using the triplet margin loss. The distillation loss $\mathcal{L}_{\text{KD}}$ follows the ranking-aware knowledge distillation formulation proposed in KDF\cite{wang2025ranking}, which preserves the relative similarity ranking among embeddings from different tasks. In our training scheme, $\mathcal{L}_{\text{PR}}$ is computed exclusively using current-domain samples within the mixed batch, whereas both $\mathcal{L}_{\text{KD}}$ and $\mathcal{L}_{\text{Rehearsal}}$ are applied to the memory samples.

Together with the loss-aware supervision introduced in Eq.~\eqref{eq:mse}, the proposed objective jointly enhances both the selection of informative samples and their effective utilization during continual learning.

\section{EXPERIMENT}
\label{sec:experiment}
\subsection{Datasets and Experimental Protocol}
\label{sec:datasets and experimental protocol}
To comprehensively evaluate the effectiveness of the proposed KDF+ framework, we conduct extensive experiments on three publicly available LPR datasets: Oxford RobotCar\cite{maddern20171}, MulRan\cite{kim2020mulran}, and the In-house dataset\cite{uy2018pointnetvlad}. Detailed statistics for all datasets are provided in Table~\ref{tab:dataset_statistics}. For the MulRan\cite{kim2020mulran} dataset, we use two distinct environments—DCC (urban) and Riverside (river-side suburban)—captured under diverse environmental conditions.

Following prior CL-LPR studies\cite{knights2022irosincloud,cui2023ralccl, liu2024micl,wang2025ranking}, we apply standard preprocessing to all point cloud frames, including ground removal and voxel downsampling.
For fair comparison across methods, we adopt the widely used \emph{4-step} training protocol:  
\[
\text{Oxford} \rightarrow \text{DCC} \rightarrow \text{Riverside} \rightarrow \text{In-house}.
\]
These four steps differ significantly in terms of location, environment, sensor configuration, and collection period, and can therefore be regarded as four distinct domains.

During training on Oxford\cite{maddern20171} and In-house\cite{uy2018pointnetvlad}, the positive/negative distance thresholds are set to 10\,m and 50\,m, respectively, while a 25\,m threshold is used during testing. For DCC\cite{kim2020mulran} and Riverside\cite{kim2020mulran}, the positive/negative thresholds are set to 10\,m and 20\,m during training, with a 10\,m threshold applied at test time.

\begin{table}[t!]
        \centering
        \caption{Statistics of the experiment datasets.}
        \label{tab:dataset_statistics}
        \setlength{\tabcolsep}{5pt}
        \normalsize
        \renewcommand\arraystretch{1.2}
        \resizebox{\columnwidth}{!}{
        \begin{threeparttable}
        \begin{tabular}{lcccc}
        \toprule
        Datasets & Oxford\cite{maddern20171} & DCC\cite{kim2020mulran} & Riverside\cite{kim2020mulran} & In-house\cite{uy2018pointnetvlad} \\
        \midrule
        Location & Oxford & Dajeon & Sejong & Singapore \\ 
        Sensor & SICK & Ouster  & Ouster & Velodyne  \\
        Collection Date & 2014-2015 & 2019 & 2019 & 2017 \\
        Training Scans & 22k & 5.5k & 5.5k & 6.7k \\
        Testing Scans & 3k & 15k & 18.6k & 1.8k \\
        \bottomrule
        \end{tabular}
        \end{threeparttable}  
        } 
\end{table} 
\begin{table*}[t!]
        \centering
        \caption{Continual learning results under the 4-step protocol. The results are presented for three different backbone architectures: PointNetVLAD\cite{uy2018pointnetvlad}, MinkLoc3D\cite{komorowski2021minkloc3d}, and CASSPR\cite{xia2023casspr}. }
        \label{tab:main_results}
        \belowrulesep=0pt
        \aboverulesep=0pt
        \normalsize
        \resizebox{0.9\textwidth}{!}{
        \begin{threeparttable}
        \begin{tabular}{lcccccccc}
        \toprule
        \multirow{2}*{Methods}&\multicolumn{2}{c}{PointNetVLAD \cite{uy2018pointnetvlad}}&\multicolumn{2}{c}{MinkLoc3D \cite{komorowski2021minkloc3d}}&\multicolumn{2}{c}{CASSPR \cite{xia2023casspr}}   &\multicolumn{2}{c}{Overall}    \\  \cmidrule(lr){2-3}\cmidrule(lr){4-5} \cmidrule(lr){6-7}\cmidrule(lr){8-9}
        ~&mR@1$\uparrow$&F$\downarrow$&mR@1$\uparrow$&F$\downarrow$&mR@1$\uparrow$&F$\downarrow$&mR@1$\uparrow$&F$\downarrow$ \\ \hline
        Fine-Tuning    & 57.97& 21.28& 72.49& 22.40&75.85 &18.88 & 68.77 & 20.85\\
        LwF \cite{li2017learning}           & 58.11& 20.49& 73.64& 19.82&78.08&15.58 & 69.94 & 18.63\\
        EWC \cite{kirkpatrick2017overcoming} & 58.21& 23.27& 78.14& 14.15&80.80&13.35 & 72.38 & 16.92\\
        InCloud \cite{knights2022irosincloud} & 63.03& 12.83& 83.66& 5.99&83.61&6.85& 76.77& 8.56\\
        CCL \cite{cui2023ralccl} & 60.59&6.51&81.47&7.11&78.4&2.53&73.49&5.38 \\
        MICL  \cite{liu2024micl} & 62.06 & 9.41 & 83.49 & 6.10 &85.20&5.02& 76.92 & 6.84 \\
        KDF \cite{wang2025ranking}    & 66.63 & 8.38& 86.40 & 1.20&85.67&1.75& 79.57& 3.93\\ 
        KDF+     & $\textbf{69.53}$ & $\textbf{2.84}$& $\textbf{86.56}$ & $\textbf{-0.01}$&$\textbf{87.30}$&$\textbf{-0.51}$& $\textbf{81.13}$& $\textbf{0.77}$\\ 
        \bottomrule
        \end{tabular}
        \begin{tablenotes}
                \item[] The best results of the experiment are shown in \textbf{bold}.
        \end{tablenotes}
        \end{threeparttable}  
        }
\end{table*}

\begin{figure}[t!]
        \centering
        \includegraphics[width=\columnwidth]{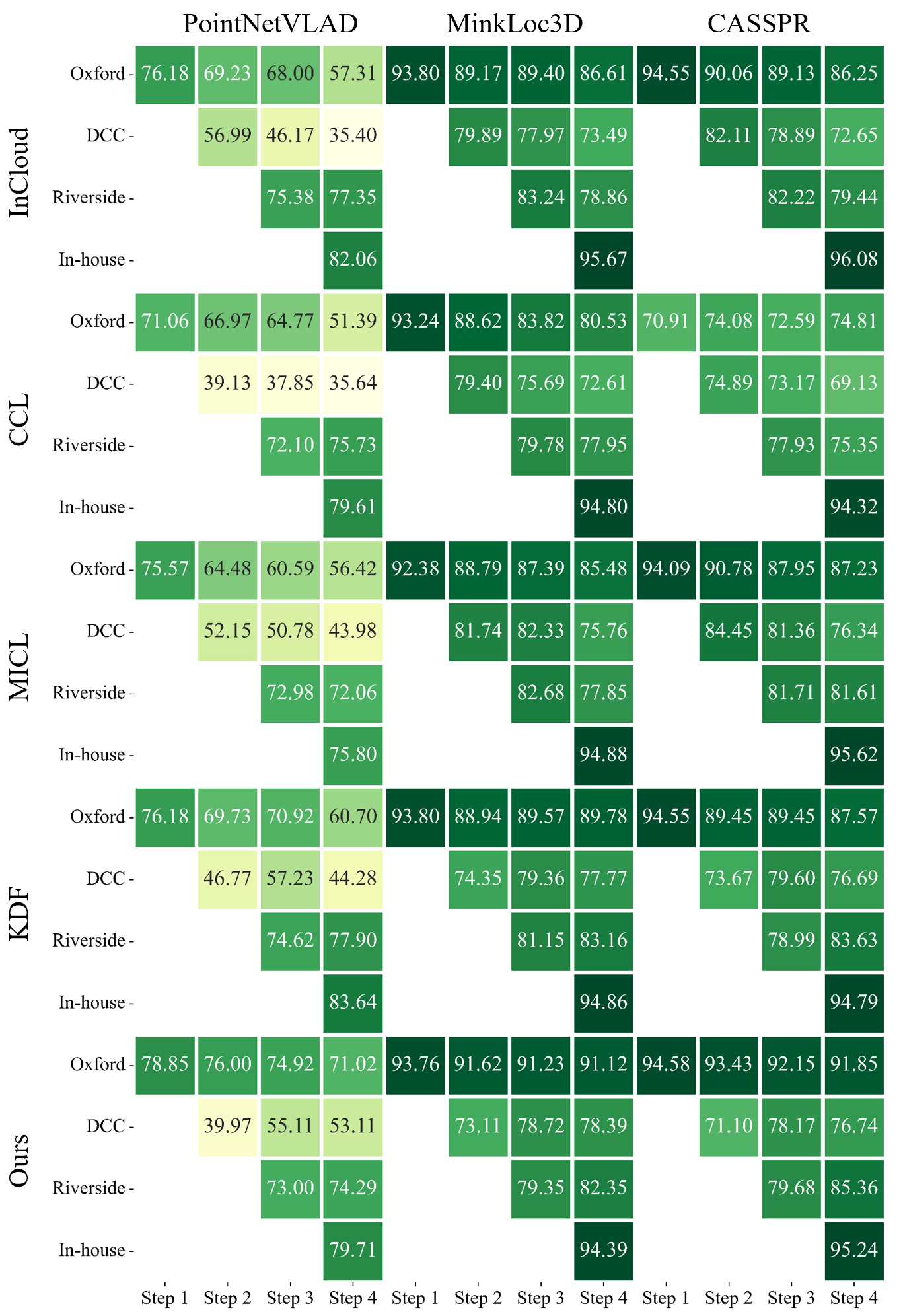}
        \caption{Detailed performance comparison under the \emph{4-step} protocol across three LPR backbones—PointNetVLAD\cite{uy2018pointnetvlad}, MinkLoc3D\cite{komorowski2021minkloc3d}, and CASSPR\cite{xia2023casspr}—and four continual learning methods: InCloud\cite{knights2022irosincloud}, CCL\cite{cui2023ralccl}, MICL\cite{liu2024micl}, and KDF\cite{wang2025ranking}. The results are presented as Recall@1 matrices, where each row corresponds to a continual learning method and each column corresponds to a different backbone. Within each matrix, Steps 1 through 4 report the Recall@1 performance on the environments encountered up to that step. Darker colors indicate higher Recall@1 values.
        }\label{fig:detailed Recall@1}
\end{figure}
\subsection{Backbones and Baselines}
To verify the generality and robustness of our approach, we evaluate KDF+ on three representative LPR backbones: PointNetVLAD\cite{uy2018pointnetvlad}, MinkLoc3D\cite{komorowski2021minkloc3d}, and CASSPR\cite{xia2023casspr}. We compare against seven baselines, including:  
(1) Fine-tuning, which serves as a lower bound without any continual learning mechanism;  
(2) LwF\cite{li2017learning} and EWC\cite{serra2018overcoming}, two classical regularization-based continual learning methods that do not use memory replay; and  
(3) four CL-LPR methods—InCloud\cite{knights2022irosincloud}, CCL\cite{cui2023ralccl}, MICL\cite{liu2024micl}, and KDF\cite{wang2025ranking}—which integrate memory replay and knowledge distillation.  
All replay-based baselines construct the memory buffer using random sampling.

\subsection{Training Details}
During the initial training stage on Oxford\cite{maddern20171}, the backbone model and the loss-aware sampling layer are jointly optimized using the triplet margin loss and the loss-aware supervision loss. Subsequent domains are learned under the continual learning setting using the proposed sampling and rehearsal strategies.

We adopt the Adam optimizer with a weight decay of $1 \times 10^{-3}$. The memory buffer size is set to 512, following prior CL-LPR works\cite{knights2022irosincloud,cui2023ralccl, liu2024micl, wang2025ranking}. Each domain is trained for 60 epochs with an initial learning rate of $1 \times 10^{-4}$, which is reduced by a factor of 0.1 after 30 epochs. The learning rate of the loss-aware layer is set to half that of the backbone network to stabilize its training.
The batch size increases progressively from 16 with a growth rate of 1.4, capped at a maximum of 256. All experiments are conducted on the same machine equipped with a single NVIDIA GeForce RTX 3090 GPU to ensure fair comparison across methods.

\subsection{Evaluation Metrics}
In CL-LPR, two standard metrics are commonly used to assess performance: the mean Recall@1 and the Forgetting Score. The mean Recall@1 measures the model's overall retrieval performance across all environments after continual training, while the Forgetting Score quantifies the average performance degradation relative to the best performance previously achieved in each environment. Formally, these metrics are defined as:
\begin{equation}
\text{mR@1} = \frac{1}{T} \sum_{t=1}^{T} R_{T,t},
\end{equation}
\begin{equation}
        F =  \frac{1}{T-1}\sum_{t = 1}^{T-1}\left( \max_{l \in \{1,\dots,t\}} R_{l,t} - R_{T,t} \right),
\end{equation}
where $T$ is the total number of training environments, and $t$ denotes the current task index. $R_{l,t}$ represents the Recall@1 of task $t$ evaluated after training step $l$. A higher mean Recall@1 and a lower Forgetting Score indicate stronger resistance to catastrophic forgetting and better continual learning performance.

\subsection{Continual Learning Performance}
The main continual learning results under the \emph{4-step} protocol are summarized in \Reftab{tab:main_results}. Our proposed KDF+ framework consistently outperforms all baselines across all backbone architectures. 

Using the MinkLoc3D\cite{komorowski2021minkloc3d} backbone, KDF+ achieves a mean Recall@1 of 86.56\% and a Forgetting Score of $-0.01$\%, surpassing the previous best method, KDF\cite{wang2025ranking}, by 0.16\% in mean Recall@1 and reducing the Forgetting Score from 1.20\% to $-0.01$\%. Similar improvements are observed with the other two backbones, PointNetVLAD\cite{uy2018pointnetvlad} and CASSPR\cite{xia2023casspr}, where KDF+ achieves gains of 2.90\% and 1.63\% in mean Recall@1, respectively. In addition, KDF+ consistently lowers the Forgetting Score, indicating not only strong retention but also enhanced performance on previously learned tasks.

These results (see \Reffig{fig:detailed Recall@1}) demonstrate the effectiveness of the proposed loss-aware sampling and rehearsal enhancement mechanisms in mitigating catastrophic forgetting and improving continual learning performance for LPR.

\begin{figure}[t!]
        \centering
        \includegraphics[width=\columnwidth]{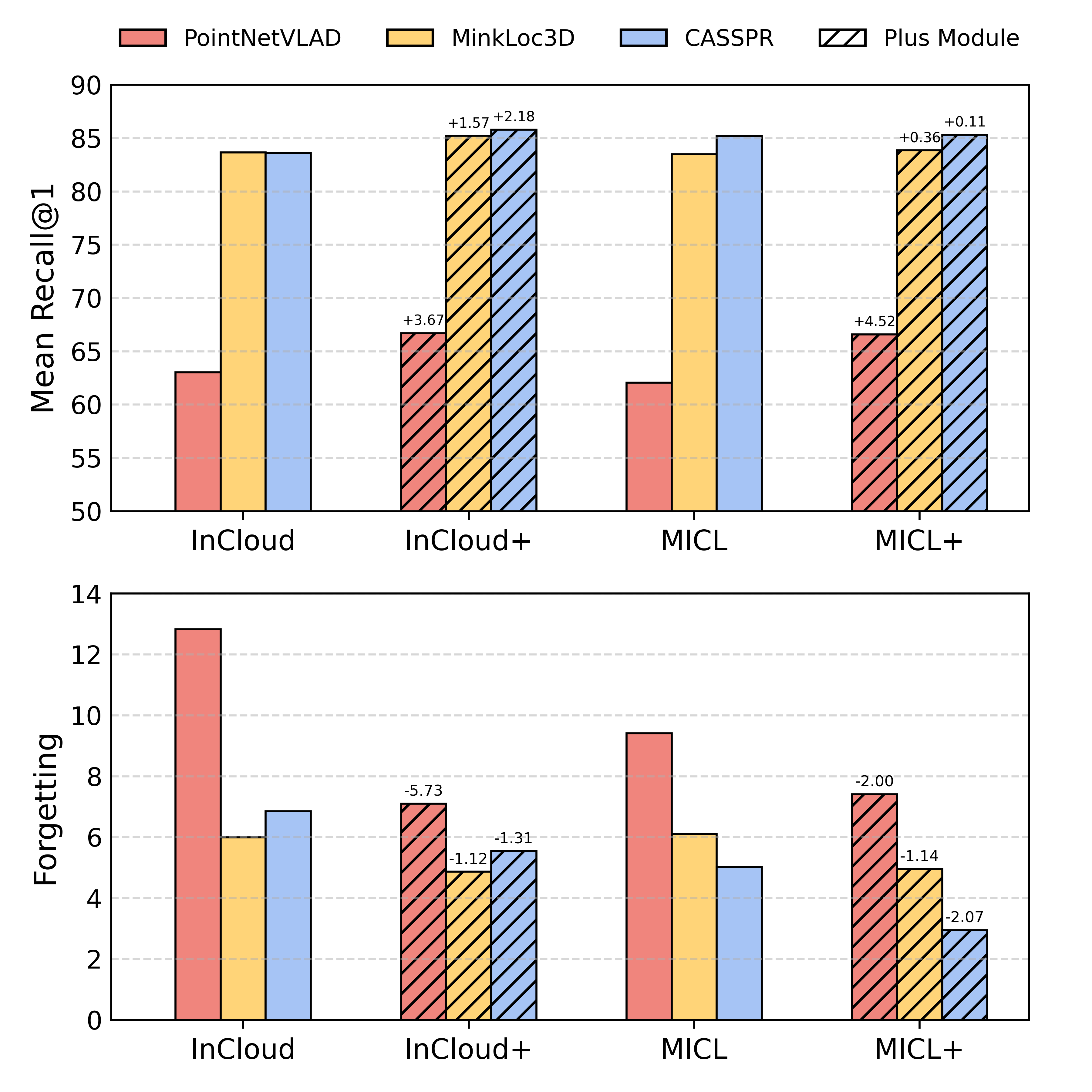}
        \caption{Generalization of the proposed KDF+ components to other CL-LPR methods, InCloud\cite{knights2022irosincloud} and MICL\cite{liu2024micl}. Integrating the loss-aware sampling and rehearsal enhancement mechanisms into these baselines yields significant improvements in both mean Recall@1 and Forgetting Score.}\label{fig:generalization}
\end{figure}

\subsection{Generalization to Other CL-LPR Methods}
We further evaluate the generalization ability of the proposed KDF+ components by integrating them into two representative CL-LPR methods: InCloud\cite{knights2022irosincloud} and MICL\cite{liu2024micl}. As shown in \Reffig{fig:generalization}, incorporating the loss-aware sampling and rehearsal enhancement mechanisms into these baselines yields substantial improvements in both mean Recall@1 and Forgetting Score.

For InCloud\cite{knights2022irosincloud}, the overall mean Recall@1 increases from 76.77\% to 79.24\%, while the Forgetting Score decreases from 8.56\% to 5.84\%. Similarly, for MICL\cite{liu2024micl}, the overall mean Recall@1 improves from 76.92\% to 78.58\%, and the Forgetting Score is reduced from 6.84\% to 5.11\%.

These results demonstrate that the proposed KDF+ components are not only effective within our own framework but also generalize well to enhance the performance of other state-of-the-art CL-LPR methods.

\begin{table}[t]
        \centering
        \caption{Sensitivity Analysis of $\omega$ on the MinkLoc3D\cite{komorowski2021minkloc3d} Backbone}
        \label{tab:ablation_w}
        \normalsize
        \resizebox{\columnwidth}{!}{
        \begin{tabular}{ccc}
        \toprule
        Rehearsal enhancement weight $\omega$ & mR@1 $\uparrow$ & Forgetting $\downarrow$ \\
        \midrule
        0.01 & 85.55 & 0.44  \\
        0.05 & 86.02 & \textbf{-0.23} \\
        0.08 & \textbf{86.56} & -0.01 \\
        0.1  & 86.38 & 0.21  \\
        0.5  & 85.79 & 0.30  \\
        \bottomrule
        \end{tabular}}
        \begin{tablenotes}
                \item[] The best results of the experiment are shown in \textbf{bold}.
        \end{tablenotes}
\end{table}
\subsection{Parameter Sensitivity Analysis}
We further analyze the sensitivity of the rehearsal enhancement weight $\omega$ in the overall loss function. As shown in Table~\ref{tab:ablation_w}, we vary $\omega$ from 0.01 to 0.5 and evaluate its impact on continual learning performance using the MinkLoc3D\cite{komorowski2021minkloc3d} backbone.

When $\omega = 0.08$, the model achieves the best performance, with a mean Recall@1 of 86.56\% and a Forgetting Score of $-0.01$\%. This indicates that an appropriate weighting of the rehearsal enhancement loss is crucial for balancing the retention of old knowledge and the learning of new information. When $\omega$ is too small (e.g., 0.01), the rehearsal enhancement effect becomes insufficient, resulting in increased forgetting. Conversely, a large $\omega$ (e.g., 0.5) places excessive emphasis on preserving old knowledge, which can hinder adaptation to new tasks.
These results highlight the importance of properly tuning $\omega$ to achieve optimal performance in continual learning.

\begin{table}[t!]
        \centering
        \caption{Ablation study of the proposed KDF+ framework on the MinkLoc3D\cite{komorowski2021minkloc3d} backbone.}
        \label{tab:ablation_study}
        \setlength{\tabcolsep}{6pt}
        \normalsize
        \resizebox{\columnwidth}{!}{
        \begin{threeparttable}
        \begin{tabular}{lcc}
        \toprule
        Methods & mR@1 $\uparrow$ & Forgetting $\downarrow$ \\ 
        \midrule
        Full KDF+ & \textbf{86.56} & \textbf{-0.01} \\
        w/o Loss-aware sampling & 86.01 &1.43 \\
        w/o Rehearsal Enhancement & 86.24 &0.39 \\
        w/o Experience Replay   & 86.02 &2.02 \\
        w/o Equal Domain   & 86.18 &0.17 \\
        \bottomrule
        \end{tabular}
        \begin{tablenotes}
                \item[] The best results of the experiment are shown in \textbf{bold}.
        \end{tablenotes}
        \end{threeparttable}
        }
\end{table}
\subsection{Ablation Study}
To evaluate the contribution of each component in the KDF+ framework, we conduct ablation experiments using the MinkLoc3D backbone. The results are summarized in Table~\ref{tab:ablation_study}. Removing either the proposed loss-aware sampling strategy (i.e., replacing it with random sampling) or the rehearsal enhancement mechanism leads to a clear degradation in performance, demonstrating the necessity and effectiveness of both components.

In addition, we analyze the impact of different replay strategies and memory balancing schemes on continual learning performance. When memory samples are directly mixed with new-task data for joint training—rather than being interleaved via experience replay—we observe a noticeable drop in accuracy, indicating the importance of structured replay. 
Likewise, if the memory buffer is not balanced across domains (e.g., when adopting the max replacement strategy from InCloud\cite{knights2022irosincloud}, which substitutes samples with those from overrepresented environments), the model exhibits further performance decline.

Overall, these findings highlight the crucial role of each design choice within KDF+, collectively contributing to its superior performance in continual LiDAR place recognition.

\section{CONCLUSION}
\label{sec:conclusion}
In this paper, we present KDF+, a novel continual learning framework for LiDAR place recognition that integrates loss-aware sampling and rehearsal enhancement mechanisms. The loss-aware sampling strategy prioritizes the selection of informative samples based on their predicted loss values, while the rehearsal enhancement mechanism encourages memory samples to further improve their performance as the model learns new tasks. 
Extensive experiments across multiple datasets demonstrate that KDF+ outperforms existing continual learning methods, achieving superior retention of previously acquired knowledge while effectively adapting to new environments. Moreover, the proposed components in KDF+ can be seamlessly incorporated into other CL-LPR frameworks, offering a promising direction for improving continual learning performance. 
In future research, we aim to investigate more advanced architectures as well as more expressive memory representations, with the goal of further enhancing knowledge retention in continual learning for LiDAR place recognition.

\bibliographystyle{IEEEtran}
\bibliography{IEEEabrv, paper}

\end{document}